\documentclass[lettersize,journal]{IEEEtran}
\usepackage{amsmath,amsfonts}
\usepackage{algorithmic}
\usepackage{algorithm}
\usepackage{array}
\usepackage[caption=false,font=normalsize,labelfont=sf,textfont=sf]{subfig}
\usepackage{textcomp}
\usepackage{stfloats}
\usepackage{xcolor}
\usepackage{url}
\usepackage{verbatim}
\usepackage{graphicx}
\usepackage{cite}
\usepackage{amsmath,amssymb,amsfonts}

\usepackage{textcomp}

\usepackage{multirow}
\usepackage{float}
\usepackage{hyperref}
\begin{document}

\title{In-depth Analysis of Privacy Threats in Federated Learning for Medical Data}

\author{Badhan Chandra Das, M. Hadi Amini, \IEEEmembership{Senior Member, IEEE}  and Yanzhao Wu, \IEEEmembership{Member, IEEE}
\thanks{ }
\thanks{Badhan Chandra Das, M. Hadi Amini, and Yanzhao Wu are  with the Knight Foundation School of Computing and Information Sciences (KFSCIS), Florida International University (FIU), Miami, FL-33199, USA. Badhan Chandra Das and M. Hadi Amini are also with the Sustainability, Optimization, and Learning for InterDependent networks laboratory (solid lab) at FIU. (E-mails: bdas004@fiu.edu, moamini@fiu.edu, yawu@fiu.edu). \\Corresponding authors: M. Hadi Amini and Yanzhao Wu.\\ This paper is an extended version of our conference paper~\cite{das2023privacy}.}}

\markboth{}
{Das \MakeLowercase{\textit{et al.}}: In-depth Analysis of Privacy Threats in Federated Learning for Medical Data}

\IEEEpubid{ }

\maketitle

\begin{abstract}

Federated learning is emerging as a promising machine learning technique in the medical field for analyzing medical images, as it is considered an effective method to safeguard sensitive patient data and comply with privacy regulations. However, recent studies have revealed that the default settings of federated learning may inadvertently expose private training data to privacy attacks. Thus, the intensity of such privacy risks and potential mitigation strategies in the medical domain remain unclear. In this paper, we make three original contributions to privacy risk analysis and mitigation in federated learning for medical data. First, we propose a holistic framework, MedPFL, for analyzing privacy risks in processing medical data in the federated learning environment and developing effective mitigation strategies for protecting privacy. Second, through our empirical analysis, we demonstrate the severe privacy risks in federated learning to process medical images, where adversaries can accurately reconstruct private medical images by performing privacy attacks. Third, we illustrate that the prevalent defense mechanism of adding random noises may not always be effective in protecting medical images against privacy attacks in federated learning, which poses unique and pressing challenges related to protecting the privacy of medical data. Furthermore, the paper discusses several unique research questions related to the privacy protection of medical data in the federated learning environment. We conduct extensive experiments on several benchmark medical image datasets to analyze and mitigate the privacy risks associated with federated learning for medical data.
\end{abstract}


\section{Introduction}
Federated Learning (FL) is an emergent Machine Learning (ML) technique where training data is distributed across multiple clients instead of a central server to protect privacy. In this approach, the training occurs locally on each client (also known as participants) and the model parameters are aggregated on a central server~\cite{FLmain1}, \cite{li2020review}. One of the most significant advantages of FL is that it can mitigate the systemic privacy risks of traditional centralized ML by keeping the private data decentralized on the clients' end and only sharing the extracted gradient updates to the central server. There are several additional benefits of FL except decentralizing the private training data including, scalability and efficiency~\cite{li2020review,zhang2022federated}. FL ensures scalability by allowing seamless integration of a large number of edge devices on clients into the learning process. It also demonstrates enhanced efficiency by collaborating among participating devices, which enables them to collectively contribute to updating the shared global model through their private training data. There are several FL algorithms prevalent such as FedAvg~\cite{FLmain1}, FedProx~\cite{li2020federated}, FedGAN~\cite{rasouli2020fedgan}, and ProxyFL~\cite{kalra2023decentralized}.

In the healthcare sector, the integration of ML and Convolutional Neural Networks (CNN) algorithms are common for the analysis of diverse medical data such as medical images, health records, and text-based doctor’s advice~\cite{qayyum2020secure}, \cite{shailaja2018machine}. These algorithms are also being utilized for prediction purposes~\cite{habehh2021machine}. Those models help to make better decision and recommendation systems in the healthcare context~\cite{sahoo2019deepreco}. Additionally, FL emerged as a promising learning technique in the medical domain due to its decentralized nature of private training data~\cite{li2020review}. It facilitates keeping the patients' sensitive health records private at their corresponding ends. FL ensures privacy-preserving ML by collaborating with multiple distributed clients, such as hospitals or clinics, without sharing sensitive raw data~\cite{nguyen2022federated}. 

Medical data, for example, X-ray images, diabetic test reports, and Magnetic Resonance Imaging (MRI) scans are considered highly sensitive records. Because those contain confidential details such as individuals' names, dates of birth, and comprehensive medical histories, which collectively serve as unique identifiers of an individual. The exposure of such sensitive information can yield severe consequences for patients, ranging from social stigma and discrimination to potential job loss and insurance coverage denial. To mitigate these risks, numerous data protection regulations have been imposed globally, underscoring the critical importance of safeguarding individuals' health-related information. The most common regulations include the Health Insurance Portability and Accountability Act (HIPAA)~\cite{cheng2006health}, the California Consumer Privacy Act (CCPA)~\cite{goldman2020introduction}, and the European Union General Data Protection Regulation (GDPR)~\cite{regulation2018general} are the most common. These regulatory boards aim to ensure the privacy and security of medical data, shielding patients from the risk of unauthorized exposure. Thus, it ensures the robust security and confidentiality of healthcare data.

Although the primary purpose of FL is to prevent leaving private training data from local devices to mitigate privacy risks, recent studies outlined that the default privacy schema in FL is inadequate to prevent privacy leakage attacks. Studies revealed that FL systems are susceptible to privacy leakage attacks where the adversaries intercept the local gradient updates transmitted by the clients before model aggregation. It can reconstruct the clients' private training data with high reconstruction accuracy, thus covertly and illegally exposing clients' privacy~\cite{geiping2020inverting,fowl2022robbing}. This vulnerability poses a severe threat to the security of FL systems, compromising the protection of client privacy \cite{zhang2021survey}. The occurrence of such privacy attacks underscores the pressing need for robust privacy-preserving mechanisms within FL frameworks to ensure the confidentiality of sensitive data. 

Protecting sensitive medical data from unauthorized access is crucial to uphold confidentiality, privacy, and the trust of patients so that legal standards can be met, and patients' privacy can be safeguarded. Despite the benefits, FL faces significant privacy vulnerabilities that pose serious threats to its application in the medical domain. Therefore, it is imperative to investigate these privacy risks and devise effective mitigation strategies to defend against privacy attacks targeting FL applications within the healthcare domain. The key contributions of this paper are as follows.

\begin{enumerate}

    \item We introduce MedPFL, a systematic framework designed for the \textbf{Med}ical Data \textbf{P}rivacy risk analysis and mitigation in \textbf{F}ederated \textbf{L}earning. The framework consists of real-world medical datasets, deep learning models, and a variety of attack and defense mechanisms. It also includes evaluation metrics for a thorough assessment of the effectiveness of different attack methods and defense strategies across various configurations.

    \item Our research highlights the significant privacy risks associated with the use of federated learning for the analysis of medical images. Through empirical evaluations, we demonstrate how adversaries can execute privacy attacks to accurately reconstruct private medical data. This finding underscores the vulnerability of FL systems to privacy breaches and the need for robust defenses.

    \item In response to the identified privacy risks, we explore various defense configurations within the FL settings. By integrating different levels of random noise, we aim to protect private medical data effectively.

    \item Our investigation reveals several challenges involved in defending against privacy attacks in FL, particularly in the context of medical data, where the urge for privacy protection is exceptionally high.

    \item We raise and discuss several research questions upon performing the experiments on different attack methods and defense mechanisms with various configurations on medical image datasets in Section \ref{section:discussion}.
  
\end{enumerate}

Systematic experiments are conducted on representative medical image datasets in order to analyze the challenges of protecting privacy for medical images in FL with visual examples of several real-world scenarios. We conjecture this investigation will capture the interest of researchers, developers, and stakeholders in the relevant domain of privacy-preserving methods in FL for medical purposes.

\section{Motivation} \label{section:motivation}

The adoption of FL within the medical domain has been accelerated, primarily to safeguard private medical data while training ML models for tasks such as COVID-19 detection using chest X-ray images~\cite{yan2021experiments}, and skin disease detection using dermoscopy images~\cite{hashmani2021adaptive}. However, recent studies have shown the inherent privacy-related challenges in FL \cite{das2023privacy, truong2021privacy, ali2022federated}. Despite numerous efforts to obscure personal health data, there remains a risk of patient information being re-identified, as evidenced by studies showcasing the potential re-identification~\cite{mileva2021risks} of individuals from DICOM images~\cite{aiello2021does}.

Furthermore, adversaries could steal the data or access the algorithm from non-encrypted networks~\cite{darzidehkalani2022federated}. 
Medical images have been reported to be susceptible to adversarial attacks due to numerous reasons, such as ambiguous ground truth \cite{finlayson2018adversarial} and highly standardized format~\cite{yoo2022open}. Therefore, the default settings of FL are still insufficient to protect the privacy of medical images. The aforementioned studies have only shown how medical images (e.g., X-ray, and MRI scans) can be leaked from the various FL environments. However, to the best of our knowledge, there is not yet a comprehensive framework for analyzing and mitigating privacy risks in FL to protect private medical data. 
Moreover, the stringent regulations on medical data protection, coupled with their distinct characteristics compared with generic data, make it imperative and much more challenging to investigate and develop effective privacy-preserving techniques for protecting medical data. For example, we highlight below several unique features of medical images. 

\begin{figure}[!t]
    \centering
    \includegraphics[width=.8\linewidth]{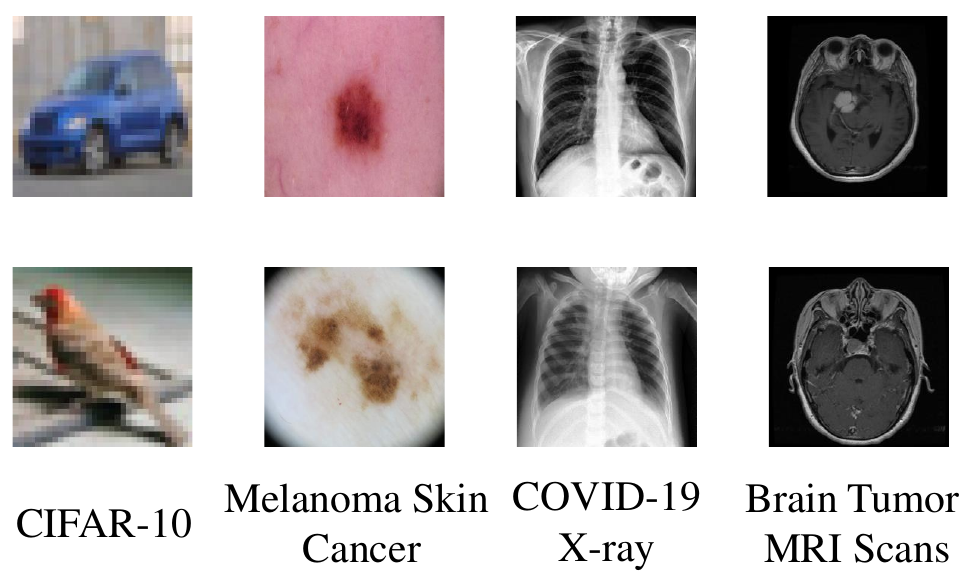}
   
    \caption{Samples of Generic Images (CIFAR-10) and Medical Images}
    \label{fig:comp1}
  
\end{figure}

\begin{enumerate}

    \item \textbf{Complexity and variability of medical images:} Medical images, such as MRI, CT scans, and ultrasound, are more complex and heterogeneous than general images (e.g., Figure~\ref{fig:comp1}). They often contain noise, artifacts, and distortions~\cite{toennies2017guide} that make data interpretation and analysis more challenging.

    \item \textbf{High dimensionality:} Medical images may have higher dimensionality. For instance, a CT scan can have hundreds of slices, each of which contains a large number of pixels. This high dimensionality requires more computational power and specialized algorithms to process, which may also impact the chance of privacy leakage.

    \item \textbf{Specificity of medical domain:} Medical images often contain specific features and structures that are not present in general images, and the interpretation of these features requires specialized knowledge and data analysis models in the medical domain.

    \item \textbf{Statistical distribution derivation:} The statistical distribution of medical images often deviates from the generic images, which creates significant differences between these two data categories in terms of processing, privacy protection, and execution time. We will discuss it in Section~\ref{section:discussion} along with experiments.

\end{enumerate}

Therefore, medical images require dedicated studies and need to be handled with more care than generic data types. In subsequent sections, we perform experiments to compare several privacy attacks to analyze privacy risks and different defense mechanisms with different configurations to prevent privacy leakage for three representative medical images in FL. Also, we demonstrate how different defense configurations impact the performance of the model. Furthermore, we distinguish different characteristics of generic images and medical images, along with several unique research challenges in terms of protecting medical data from privacy attacks.

\section{Related Works}\label{section:related-work}

 Though FL was proposed to protect clients' private data, researchers have shown that the FL is prone to be attacked by adversaries from both security and privacy perspectives \cite{mothukuri2021survey} \cite{lyu2022privacy}.

\subsection{Security Attacks}

In FL, a malicious user or an adversary takes advantage of the vulnerabilities \cite{men2019finding} and gains control of one or more participants (i.e., clients) within the FL environment so that it can cause the malfunction of the whole system \cite{mothukuri2021survey}. 

Poisoning attacks are the most common types of security attacks~\cite{munoz2017towards}. The basic concept of poisoning attacks in FL refers to a scenario where a malicious user from the participants in the FL inserts poisonous data samples or parameters intending to malfunction the whole system. Poisoning can happen in both data level~\cite{tolpegin2020data} and model level~\cite{bhagoji2018model}. For data poisoning attacks in FL, malicious clients inject mislabeled, corrupted, or poisoned data into their local training data and attempt to update the global model with the poisoned data~\cite{fung2018mitigating}. That eventually yields degraded performance of the system \cite{tolpegin2020data}, which is the primary goal of the data poisoning attack. On the other hand, in poisoning attacks, rather than modifying the training data directly, the attacker/malicious client intentionally manipulates the gradient updates before sending them to the central server~\cite{bhagoji2019analyzing}. Backdoor attacks are also prevalent in the FL setting, where the attacker aims to inject a desired malicious task into the existing model~\cite{mothukuri2021survey}.

\subsection{Privacy Attacks}

The privacy attacks in FL primarily focused on inferring sensitive information about participants' private training data based on the gradients they send to the central server \cite{mothukuri2021survey}. Training data leakage through reconstruction attacks is the most prevalent in this category. 

Recent studies~\cite{zhu2019deep,geiping2020inverting,wei2020framework,wei2021gradient,liu2022threats,dahlgaard2022analysing,wei2023securing,wei2023model-cloaking,zhao2020idlg} have demonstrated that the clients' private training data in FL environment can be reconstructed by the adversary with high reconstruction accuracy. Several privacy leakage attack techniques have been proposed including Client Privacy Leakage (CPL)~\cite{wei2020framework}, Deep Leakage from Gradients (DLG)~\cite{zhu2019deep}, Improved DLG (iDLG) \cite{zhao2020idlg}, and Inverting Gradients (GradInv)~\cite{geiping2020inverting}. These attack methods illustrate that the default privacy scheme in FL might not offer adequate protection against privacy leakage attacks in the default FL environment. These attack methods underscore the need for enhanced privacy-preserving mechanisms in FL to effectively safeguard participants' sensitive data. Random noise insertion, e.g., Gaussian noise~\cite{abadi2016deep} or Laplacian noise~\cite{melis2015efficient} might be an effective method for defending against privacy attacks in deep learning models. However, there is a significant lack of comprehensive and systematic studies addressing the potential threats posed by these privacy attacks to medical data within FL settings.

\begin{figure*}[t]
    \centering
    \includegraphics[scale=.35]{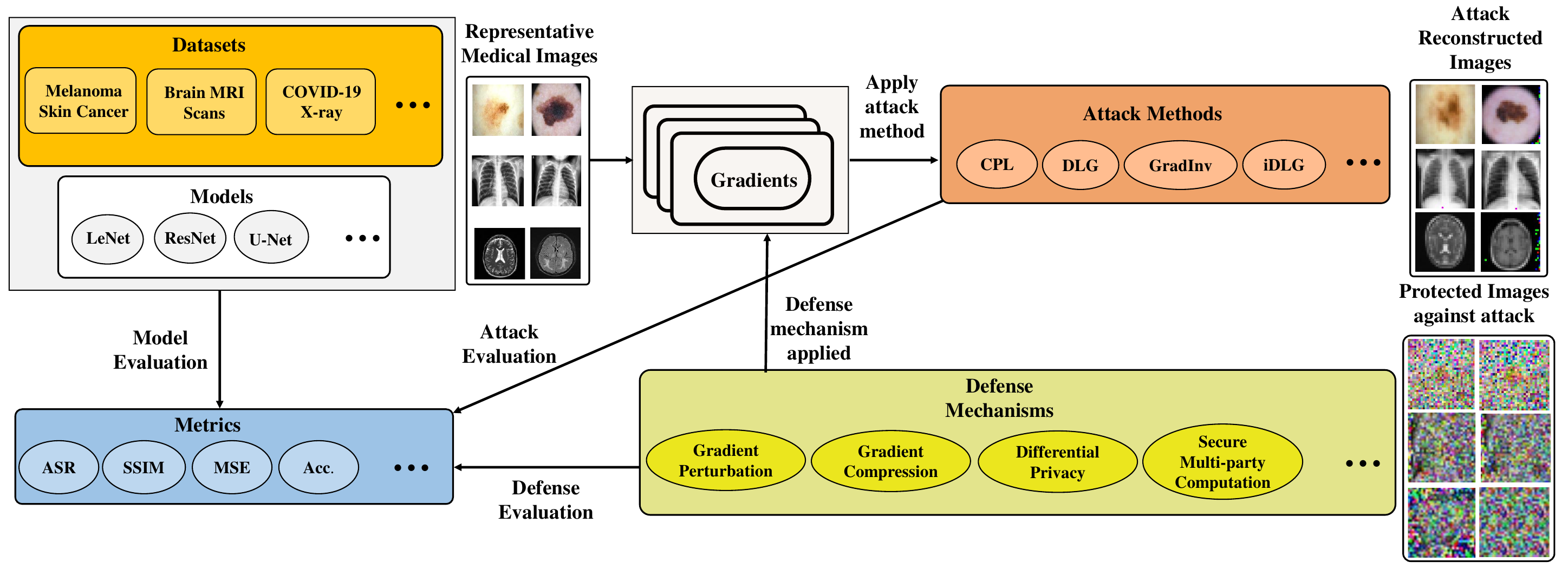}
 
    \caption{Overview of MedPFL: a framework for \textbf{Med}ical Data \textbf{P}rivacy risk analysis and mitigation in \textbf{F}ederated \textbf{L}earning}
    \label{fig:framework}
   
\end{figure*}
In the medical domain, Kaissis et al.~\cite{kaissis2020secure} surveyed privacy-preservation techniques, which are designed for classifying chest X-rays and segmenting CT scans in deep learning training~\cite{kaissis2021end}. A personalized local differential privacy in the FL scheme was illustrated by Shen et al.~\cite{shen2023pldp} on the MNIST dataset to overcome the challenges, (e.g., inadequate or excessive privacy protection due to the same privacy budget) of the existing local differential privacy-based FL scheme. The use of Differential Privacy (DP) to protect medical data from such privacy attacks has been discussed by Liu et al.~\cite{liu2023survey}. Adnan et al.~\cite{adnan2022federated} proposed DP in FL on histopathology images, but they did not explicitly mention any vulnerabilities or unique challenges of medical images in a decentralized environment. Aouedi et al.~\cite{aouedi2022handling} highlighted the several challenges in FL, focusing on privacy and security concerns, issues related to client synchronization, and the complexity arising from the presence of non-IID datasets.

We propose a comprehensive framework to specifically analyze the privacy risks inherent in medical data and their mitigation strategies in the FL environments. Moreover, our research has identified a set of unique challenges and distinguishing features, notably the intricate nature, higher-dimensional aspects, and latent pathological information inherent in medical images. These factors significantly amplify privacy concerns surrounding medical data in FL scenarios~\cite{yoo2022open}. Therefore, the extent of vulnerability of FL applications in the medical sector to privacy attacks lacks in-depth studies, along with the optimal approaches for mitigating such risks. This poses critical challenges in employing FL for processing sensitive private medical data, such as skin cancer images, X-ray images, and MRI scans of patients. Addressing these gaps in understanding is imperative for ensuring the security and privacy of sensitive medical data within FL frameworks.

\section{Framework Overview} \label{section:framework-overview}
We propose MedPFL, a comprehensive framework that addresses the critical need for privacy risk analysis and mitigation in FL, particularly in the medical domain. It comprises five key components that aim to streamline the evaluation, comparison, and mitigation of privacy risks associated with processing medical data within FL environments. Figure~\ref{fig:framework} illustrates these components and major workflows for privacy attacks on the trained model with private medical datasets, and defense mechanisms for safeguarding the private medical images from being attacked. We also incorporate several evaluation metrics to measure the efficacy of both attack and defense mechanisms.

\subsection{Datasets}
The proposed framework offers a collection of real-world medical datasets from publicly available sources. These datasets represent diverse forms of medical data, including Melanoma Skin Cancer images~\cite{skin}, COVID-19 X-ray images~\cite{covid}, and Brain Tumor MRI scans~\cite{brain_MRI}. These datasets serve as valuable resources to assess the potential privacy risks associated with medical data. Additionally, our framework provides easy-to-use APIs to facilitate the integration of new datasets.
The details of the datasets that we used in this study will be introduced later in Section~\ref{section:experimental-analysis} (experimental analysis).

\subsection{Models}

Our framework supports a variety of deep-learning models for medical data processing. For instance, we incorporate LeNet~\cite{lecun1998gradient} and ResNet~\cite{he2016deep} for medical image classification tasks, and U-Net~ \cite{ronneberger2015u} for biomedical image segmentation. These models represent the mainstream Deep Neural Network (DNN) architectures for medical data analysis. It allows us to reflect the research challenges by scrutinizing their privacy vulnerabilities and investigating potential mitigation techniques. For now, the framework supports classification tasks, in the future, we are planning to incorporate other learning tasks, e.g., medical image segmentation.

\vspace{-1.5ex}
\subsection{Attack Methods}

To assess the privacy risks associated with medical data, we implemented a set of attack methods within our framework. These include CPL~\cite{wei2020framework}, DLG~\cite{zhu2019deep}, iDLG~\cite{zhao2020idlg}, and GradInv~\cite{geiping2020inverting}. Leveraging these attack techniques, we conducted privacy attacks against diverse medical datasets and models to evaluate their potential privacy risks. Evaluation of these privacy attack methods is conducted using well-established metrics such as Attack Success Rate (ASR), Mean Squared Error (MSE), and Structural Similarity Index Measure (SSIM).

\vspace{-2ex}
\subsection{Defense Mechanisms}

In a conventional FL environment, several defense mechanisms have been proposed to prevent various types of privacy attacks. These defenses include gradient perturbation~\cite{wei2020framework}, gradient compression~\cite{wei2020framework}, secure multi-party computation~\cite{goldreich1998secure}, and DP techniques~\cite{abadi2016deep}. However, there is a lack of comprehensive studies on the efficacy and optimal configurations of these defense mechanisms against a wide range of privacy attacks within the realm of medical data. In response to this gap, our MedPFL framework offers a range of defense mechanisms, facilitating in-depth investigations into their effectiveness and the factors influencing their performance against privacy attacks on medical data. Additionally, we provide a set of evaluation metrics designed to assess the efficacy of these defense mechanisms in safeguarding the privacy of medical data.

\vspace{-3ex}
\subsection{Evaluation Metrics}
Here, we briefly introduce three major evaluation metrics that we used in this study.

\textbf{Attack Success Rate (ASR)} is the percentage of the number of successfully reconstructed samples over the number of samples attacked~\cite{wei2020framework}. This metric can be used to evaluate the performance of various attack methods and defense mechanisms. A higher ASR value indicates the high efficacy of a privacy attack method and lower ASR values refer to better performance of the defense mechanisms.

\textbf{Mean Squared Error (MSE)} is used to quantify the average squared difference between the pixel values of two images, providing a numerical measure of the dissimilarity or error between them~\cite{tan2013perceptually}. 
Lower MSE implies higher image similarity, implying smaller average pixel intensity differences, while higher MSE implies greater dissimilarity with larger average differences. MSE is used for evaluating both attack and defense mechanisms in this study. Since MSE refers to the difference between the original private image and the attack reconstructed one, for the attack methods, the lower MSE values indicate the attack is more successful, and exactly the opposite for the defense.

\textbf{Structural Similarity Index Measure (SSIM)} provides a measure of the structural similarity between two images, taking into account not only pixel intensity differences but also spatial information and human visual perception~\cite{wang2004image}. It ranges from 0 to 1 with higher values implying greater image similarity. SSIM is used for evaluating both attack and defense mechanisms. Here, SSIM values present the similarity between the original private image and the attack reconstructed one. So, for the attack methods, higher SSIM values indicate the attack is more successful, and exactly the opposite for defense.

\begin{algorithm}
\caption{\emph{FedSGD}. \emph{\#} of clients $C$, learning rate $\eta$,  \emph{\#} of local epoch $E$, $n_k$ is the number of data samples associated with client $k$ from set of $P_k$, gradient computed by each client $k$, $g_k$ \cite{FedSGD}.
  }
\begin{algorithmic}[1]
\STATE \textbf{Server's execution:}
\STATE Initialize $\omega_0$
\STATE $k \leftarrow $(random set of clients from $C$)
    \FOR{iterations  $t = 1, 2, ...$}
        \STATE \textbf{ClientUpdate($k, \omega_t$)}

    \STATE $\omega_{t+1} \leftarrow \omega_t - \eta \sum_{k=1}^{k} \frac{n_k}{n} g_k$,   $\Bigl[\sum_{k=1}^{k} \frac{n_k}{n} g_k=\nabla L(w_t)\Bigr]$
    \ENDFOR
    \STATE \textbf{Clients' execution:}
        \STATE \textbf{ClientUpdate($k, \omega_t$):}
        
        \STATE $g_k \leftarrow \nabla L_k (\omega_t)$
        \STATE Return $g_k$ to the central server.
\end{algorithmic}
\label{algm:FedSGD}
\end{algorithm}

\section{Methodology} \label{section:methodology}

In the healthcare system, FL involves the decentralization of ML models from a central server to be distributed across a group of hospitals and clinics, referred to as client nodes. Since it is uncertain that all the clients are available, a small number is chosen from the pool of participants to participate in collaborative learning during each iteration. \vspace{-3ex}

\subsection{FL Architecture}
We present the representative FL aggregation algorithm, FedSGD~\cite{FedSGD}, in Algorithm \ref{algm:FedSGD}. First, the central server initiates the global model $\omega_0$ and shares it with the selected clients at round $t$. Each client $k$ performs training and computes the gradient $g_k$ of the loss function with respect to model weights on their private training data at iteration $t$ and sends $g_k$ back to the central server. Then, the central server aggregates the gradients from the selected participating clients $k$, weighted by the number of data samples associated with $k$.

\subsection{FL Architecture in Medical Domain}
 
In the healthcare sector, FL entails duplicating ML models from a central server and disseminating them among a set of clients, including clinics, hospitals, and healthcare organizations. The procedure typically works as follows. Initially, in \textit{step 1}, each client receives a global model denoted as $\omega_0$ at round $t$ from the trusted centralized server for client $k$. In \textit{step 2}, each client updates the local model and computes gradient, $\nabla L_k (\omega_t)$, utilizing its private medical data, $T(k, t)$. Subsequently, in \textit{step 3}, gradients, $\nabla L_k (\omega_t)$, are transmitted to the central server. After that, the central server aggregates these local model updates (gradients) received from $k$ clients and adjusts its global model, often employing an aggregation technique such as FedSGD~\cite{FedSGD}. This iterative process continues until satisfying predetermined stopping criteria, such as reaching a specified number of iterations or achieving a desired level of accuracy. 

\subsection{Attack Method on Medical Images}

\begin{figure*} 

\begin{center}

    \subfloat[Attack]{\includegraphics[height=3.2cm]
    {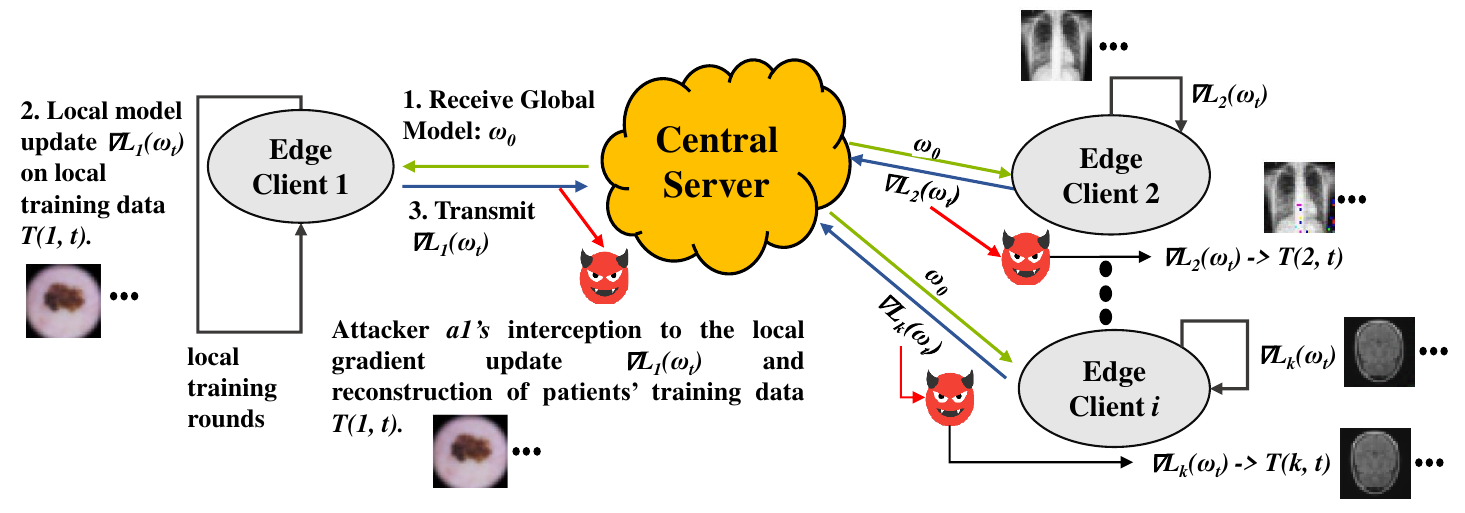}}
    \label{fig:attack_method} 
		\hspace{.00001mm}
    \subfloat[Defense]{\includegraphics[height=3.2cm]
    {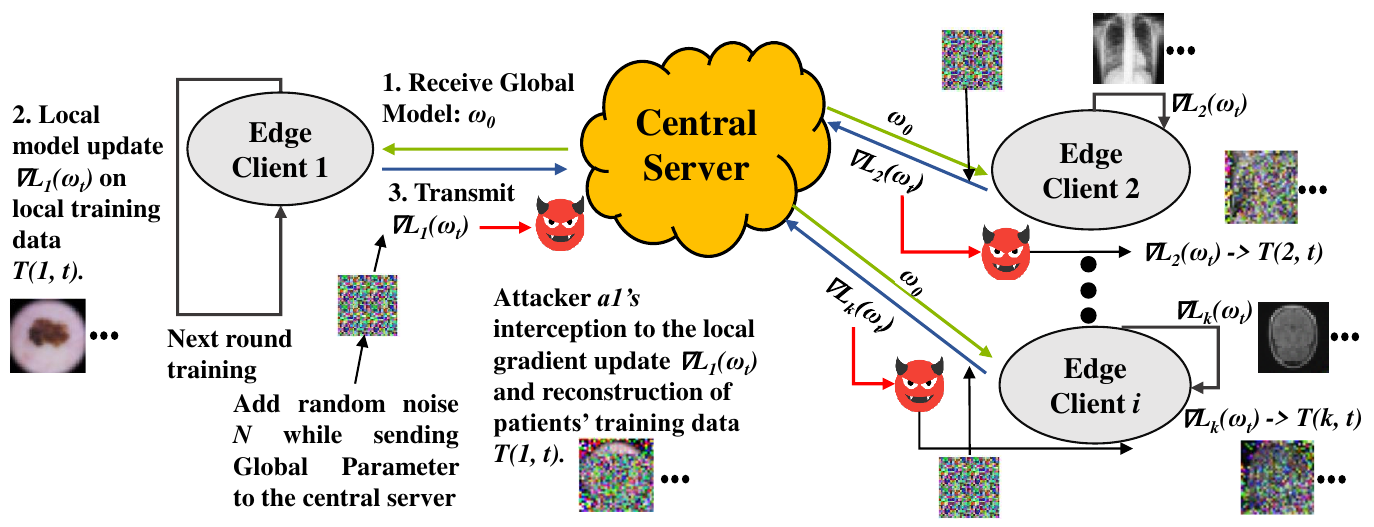}}

    \vspace{-2ex}
    \label{fig:defense_mech} 
\end{center}
\caption{Overview of Attack Method  and Defense Mechanism}
\label{fig:attack_defense_mech} 
\vspace{-1ex}
\end{figure*}

In Figure \ref{fig:attack_defense_mech}(a), we illustrate how an adversary could intercept the local gradient update and reconstruct patients' sensitive medical information. During \textit{step 3}, when the local model update (gradients) $\nabla L_k (\omega_t)$ is transmitted to the central server, an attacker $a_i$ could intercept this transmission and obtain the local model update for the respective client $k$. Subsequently, the attacker could scrutinize the periodic local model updates to execute privacy attacks, as documented in prior works such as~\cite{wei2020framework,zhu2019deep}, potentially leading to the reconstruction of client $k$'s private data. Typically, the privacy attack method starts by initiating dummy data and labels (e.g., $x', y'$) of the same size as the private training data. Then these “dummy data” are fed into the models and get “dummy gradients”. \vspace{-3ex}

\begin{equation}
    \nabla \omega'=\frac{\partial L( F(x', \omega), y')}{\partial \omega}
\end{equation}
Optimizing the dummy gradients close to the original gradients on the private data also makes the dummy data close to the real
private data. Given gradients at a certain step, we obtain the training
data by minimizing the loss function as follows.
\begin{equation}
\begin{aligned}
   x', y' &=  \arg \min_{x', y'}||\nabla \omega' - \nabla \omega||^2 \\ 
   &= \arg \min_{x', y'}||\frac{\partial L(F(x', \omega), y' )}{\partial \omega}  - \nabla \omega ||^2
\end{aligned}
\end{equation}

The attacker performs the privacy attacks by taking the dummy data, dummy label, and the local model updates (gradients obtained from intercepting \textit{step 3}) of the client from local training. Additionally, the attacker utilizes the shared global model so that it can iteratively update the dummy data and labels to reconstruct the client's private training data. This iterative process involves updating the dummy data and labels to minimize the disparity between the local gradients computed on the private data and the dummy gradients computed on the dummy data and labels, denoted as $||\nabla \omega' - \nabla \omega||^2$. This process, facilitated by the shared global model, gradually aligns the dummy data with the private training data, exacerbating privacy breaches.

\subsection{Defense Mechanism for Medical Images}

In the vanilla FL context, various methods exist to thwart privacy leakage attacks, which could potentially safeguard medical data privacy. One such method is gradient perturbation~\cite{wei2020framework}, which entails injecting a controlled amount of Laplacian or Gaussian noise into the local model update 
$\nabla L_k (\omega_t)$ during \textit{step 3} (refer to Figure~\ref{fig:attack_defense_mech}(b)). By introducing noise to $\nabla L_k (\omega_t)$, this approach introduces uncertainty into local updates, obscuring details and hindering adversaries from accurately reconstructing private data, such as medical images.

DP can be leveraged to provide theoretical guarantees in gradient perturbation to allow the sharing of confidential data and safeguard the privacy of the individuals whose data is being utilized~\cite{cynthia2006differential}. DP-based mechanisms introduce a controlled amount of noise to the private data in a manner that preserves the statistical characteristics and obscures the actual values of individual data points. By doing so, DP prevents malicious entities from identifying specific data points, thereby ensuring the privacy of the individuals concerned. As represented in Figure~\ref{fig:attack_defense_mech}(b), when we add a controlled amount of noise to the gradient at \textit{step 3} of the process, it can prevent accurate reconstruction of private training data.
Also, gradient compression~\cite{wei2020framework} can be utilized to defend against privacy leakage attacks in FL. 
Secure multi-party computation is another category of defense mechanisms where multiple parties perform computations without revealing their sensitive data to each other~\cite{goldreich1998secure}.

\section{Experimental Analysis}~\label{section:experimental-analysis}
The experiments are conducted on a GPU server with an NVIDIA RTX A6000 with 48 GB memory.

\begin{table}[!h]

\centering
\vspace{-1.8ex}
\caption{Dataset Information and Properties}
\scalebox{.725}{
\small
\begin{tabular}{|c|c|c|}
\hline
   \textbf{Dataset}             & \textbf{\# of Samples} & \textbf{\# of Classes and Names}           \\ \hline
 Melanoma Skin Cancer & 10000                   & \begin{tabular}[c]{@{}c@{}}2 \\ (Benign and malignant)\end{tabular}                   \\ \hline
 COVID-19 X-ray            & 317                    & \begin{tabular}[c]{@{}c@{}}3 \\ (COVID, Normal, Viral Pneumonia)\end{tabular}         \\ \hline
 Brain Tumor MRI Images        & 7022                   & \begin{tabular}[c]{@{}c@{}}4\\ (Pituitary, Glioma, Meningioma, No tumor)\end{tabular} \\ \hline
\end{tabular}
}
\label{tab:dataset_info}
\end{table}

\vspace{-2ex}
\subsection{Dataset Properties and Preprocessing}

We conduct experiments employing various attack methods on three representative medical image datasets: Melanoma Skin Cancer~\cite{skin}, COVID-19 X-ray~\cite{covid}, and Brain Tumor MRI Images~\cite{brain_MRI}. These datasets contain 10,000, 317, and 7,022 samples respectively, distributed across two, three, and four classes as outlined in Table~\ref{tab:dataset_info}. Each dataset comprises images of different sizes, for instance, the Melanoma Skin Cancer dataset contains images of size 300$\times$300. Also, COVID-19 X-ray images and Brain Tumor MRI Images exhibit varying shapes. In our experiment, all images were resized to 32$\times$32 and we performed data normalization using mean and standard deviation during our preprocessing phase. We employ a 4-layer CNN architecture, incorporating a fully connected layer to process input images with three channels, as outlined in CPL~\cite{wei2020framework}, DLG~\cite{zhu2019deep}, and iDLG~\cite{zhao2020idlg} respectively on the medical image datasets. Additionally, we conduct GradInv attacks~\cite{geiping2020inverting} on all three datasets to compare the attack performance with CPL, DLG, and iDLG respectively. The GradInv \cite{geiping2020inverting} attack is designed to reconstruct private training images from both ResNet-18~\cite{he2016deep} which is pre-trained on ImageNet~\cite{deng2009imagenet} and its untrained version.

\begin{figure}
    \centering
    \includegraphics[scale=0.45]{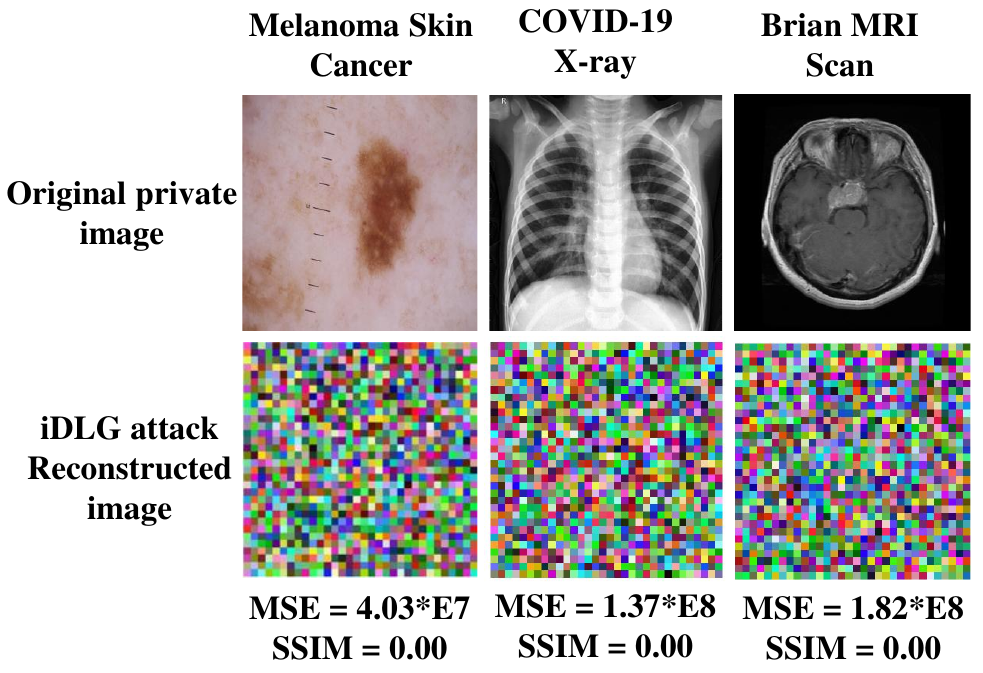}
    \vspace{-2ex}
    \caption{Samples of attack failure cases of all datasets.}
    \label{fig:revised_MSE_SSIM}
  
\end{figure}

\subsection{Attack Method Configuration}

CPL~\cite{wei2020framework}, DLG~\cite{zhu2019deep}, and iDLG~\cite{zhao2020idlg} follow a similar approach in order to reconstruct client's private data. The attacker intercepts the local model update $\nabla L_k (\omega_t)$ corresponding to a client $k$ at iteration $t$. They initialize a dummy image of the same dimensions as the training data, along with a dummy label. The dummy image is then iteratively updated to minimize the $L2$ distance between the actual gradient $\nabla L_k (\omega_t)$ computed on the private training data and the dummy gradients computed on the dummy data. In other words, the attack aims to find a dummy data sample that produces gradients as close as possible to the intercepted gradients from the client's local update. For executing the privacy attack, we employ the L-BFGS optimization method, as suggested by CPL~\cite{wei2020framework}, DLG~\cite{zhu2019deep}, and iDLG~\cite{zhao2020idlg}. The GradInv~\cite{geiping2020inverting} attack is basically conducted by leveraging the gradients, $\nabla L_k (\omega_t)$, of the local training data to reconstruct the original images utilizing a network composed of fully-connected layers. This attack iteratively analyzes the $\nabla L_k (\omega_t)$ under the condition of non-zero gradients, optimizing the angle-based loss function (cosine similarity), to reconstruct the private data. Adam is employed as the optimization algorithm. \vspace{-2ex}

\subsection{Performance of Attack Methods}

Here, we present the performance comparison of the four attack methods that we investigated in this research (i.e., CPL, DLG, iDLG, and GradInv) on different medical image datasets in Table~\ref{tab:exp_results2}. We chose 100 randomly sampled images for all the attack methods for all three datasets. During the ASR computation, if the SSIM between the original private image and the attack-reconstructed image is above or equal to $0.9$, we consider that case as a successful attack. In our experiment, we found very high MSE values and very low SSIM values between the original training image and the reconstructed one for several attack failure cases in various attack methods. We show some of the samples from the iDLG attack method in Figure \ref{fig:revised_MSE_SSIM} along with their corresponding high MSE and low SSIM values. Such high MSE and low SSIM values, i.e., attack failure cases, might cause bias in the attack performance evaluation. Therefore, we considered the average SSIM and MSE values shown in Table \ref{tab:exp_results2}, which has been calculated only for successful attacks as per ASR requirements for all attack methods. \vspace{-2ex}

\begin{table*}[!ht]

\caption{Performance Comparison of GradInv., CPL, DLG, and iDLG attack methods on all three medical datasets.}

\centering
\scalebox{1}
{
\begin{tabular}{|l|c|c|c|c|c|c|}
\hline
Dataset                                                                          & Method                    & Model                                                                     & \begin{tabular}[c]{@{}c@{}}Attack Success\\ Rate\end{tabular} & \begin{tabular}[c]{@{}c@{}}Avg.\\ SSIM for\\ Successful Attacks\end{tabular} & \begin{tabular}[c]{@{}c@{}}Avg.\\ MSE for\\ Successful Attacks\end{tabular} & \begin{tabular}[c]{@{}c@{}}Avg. Attack\\ Execution Time\\ per image (in seconds)\end{tabular} \\ \hline
\multirow{4}{*}{\begin{tabular}[c]{@{}l@{}}Melanoma \\Skin\\ Cancer\\ Dataset\end{tabular}} & \multirow{2}{*}{GradInv.} & ResNet-18 Untrained                                                       & \textbf{{76\%}}                                                 & {0.9473}                                     & {0.5301}                                    & 5940.94                                                                        \\ \cline{3-7} 
                                                                                 &                           & \begin{tabular}[c]{@{}c@{}}ResNet-18 Trained
                                                                                 \end{tabular} & 72\%                                                          & 0.9389                                              & 0.5621                                             & 4880.58                                                                        \\ \cline{2-7} 
                                                                                 & CPL                       & CNN                                                                       & 49\%                                                          & \textbf{0.9996}                                                & \textbf{4.1 * E-7}                                             & 50.8471                                                                        \\ \cline{2-7} 
                                                                                 & DLG                       & CNN                                                                       & 47\%                                                          & 0.9569                                                & 6.28 * E-3                                             & 60.1282   
                                                                                  \\ \cline{2-7} 
                                                                                 & iDLG                       & CNN                                                                       & 47\%                                                          & 0.9667                                                & 2.83 * E-4                                             & 65.6132 \\ \hline
\multirow{4}{*}{\begin{tabular}[c]{@{}l@{}}Covid-19\\ X-ray\\ Dataset\end{tabular}}      & \multirow{2}{*}{GradInv.} & ResNet-18 Untrained                                                       & \textbf{{75\%}}                                                 & {0.9305}                                     & {0.6763}                                    & 5908.71                                                                        \\ \cline{3-7} 
                                                                                 &                           & \begin{tabular}[c]{@{}c@{}}ResNet-18 Trained
                                                                                 \end{tabular} & 30\%                                                          & 0.9252                                              & 0.6954                                             & 4967.69                                                                        \\ \cline{2-7} 
                                                                                 & CPL                       & CNN                                                                       & 55\%                                                          & \textbf{0.9999}                                                & \textbf{2.88 * E-7}                                             & 92.2457                                                                        \\ \cline{2-7} 
                                                                                 & DLG                       & CNN                                                                       & 55\%                                                          & 0.9785                                                & 3.39 * E-3                                             & 125.6201      
                                                                                 \\ \cline{2-7} 
                                                                                 & iDLG                       & CNN                                                                       & 56\%                                                          & 0.9957                                                & 1.42 * E-4                                             & 81.4562  
                                                                                 \\ \hline
\multirow{4}{*}{\begin{tabular}[c]{@{}l@{}}Brain\\Tumor\\ MRI\\ Dataset\end{tabular}}   & \multirow{2}{*}{GradInv.} & ResNet-18 Untrained                                                       & {73\%}                                                 & {0.9491}                                     & {0.6061}                                    & 5953.77                                                                        \\ \cline{3-7} 
                                                                                 &                           & \begin{tabular}[c]{@{}c@{}}ResNet-18 Trained
                                                                                 \end{tabular} & 13\%                                                          & 0.9168                                              & 0.8728                                              & 4933.24                                                                        \\ \cline{2-7} 
                                                                                 & CPL                       & CNN                                                                       & \textbf{76\%}                                                          & \textbf{0.9999}                                                & \textbf{4.57 * E-7}                                             & 103.126                                                                         \\ \cline{2-7} 
                                                                                 & DLG                       & CNN                                                                       & 71\%                                                          & 0.9445                                                & 1.51 * E-2                                             & 116.485                                       \\ \cline{2-7} 
                                                                                 & iDLG                       & CNN                                                                       & 72\%                                                          & 0.9972                                                & 3.95 * E-5                                             & 75.7268                                       \\ \hline
\end{tabular}

}
\label{tab:exp_results2}

\end{table*}

\begin{figure}[h]
    \centering
    
    \includegraphics[scale=0.5]{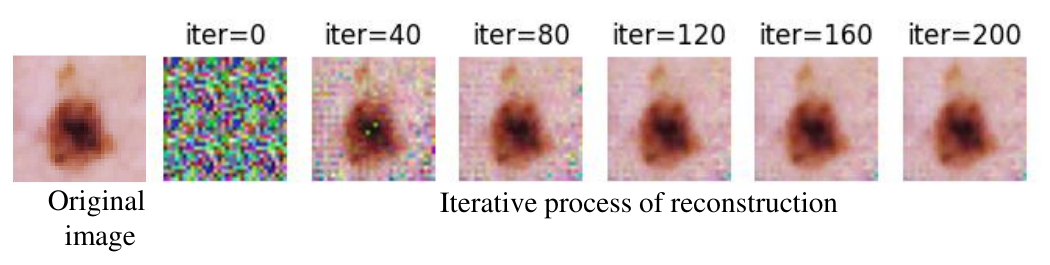}
    \vspace{-4ex}
    \caption{CPL attack on Melanoma Skin Cancer Dataset}
    \label{fig:Attack_result1}
    
    \includegraphics[scale=0.5]{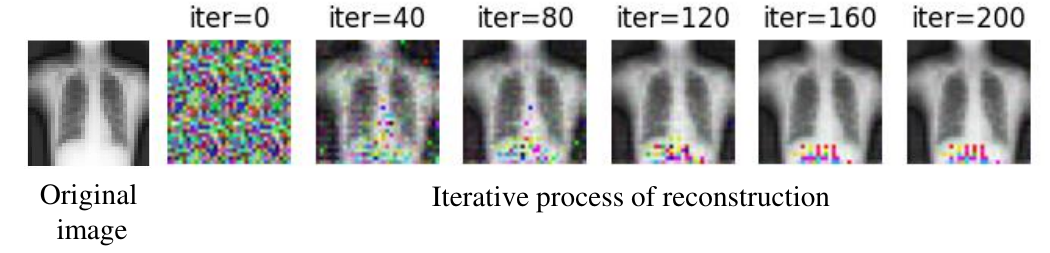}
    \vspace{-4ex}
    \caption{CPL attack on COVID-19 X-ray Dataset}
    
    \label{fig:Attack_result2}
    \vspace{0.32ex}
    \includegraphics[scale=0.5]{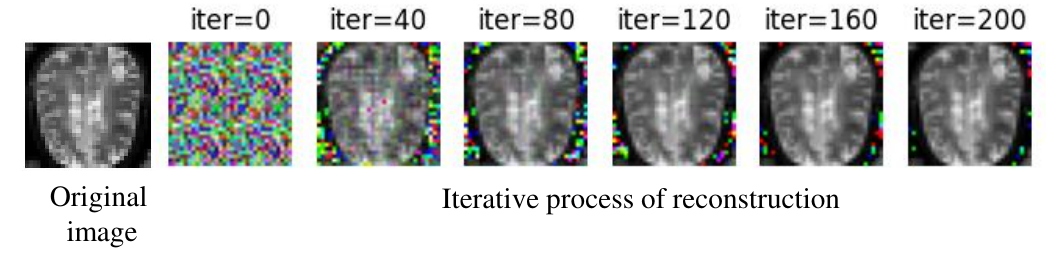}
    \vspace{-4ex}
    \caption{CPL attack on Brain Tumor MRI Images Dataset}
    
    \label{fig:Attack_result3}

\end{figure}
\begin{figure}[h]
    \centering
    \includegraphics[width=1.0\linewidth]{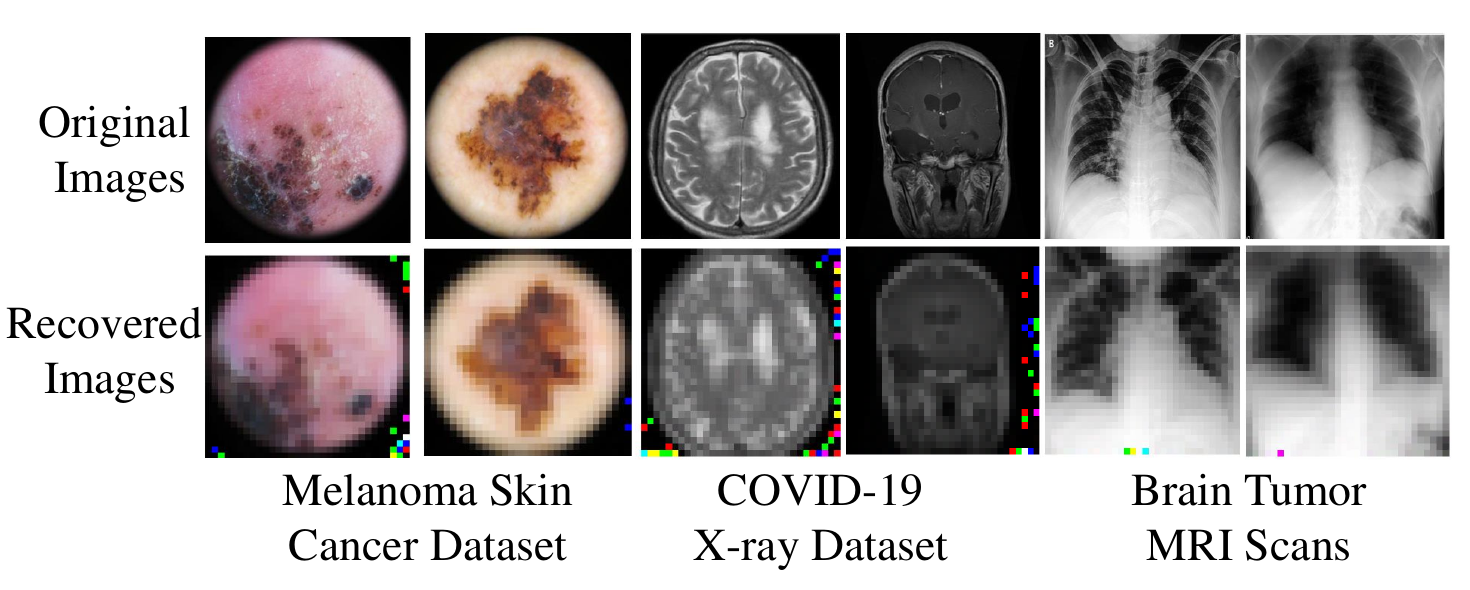}
    \vspace{-4ex}
    \caption{iDLG attack on Medical Image Datasets (Top row: original private training images, Bottom Row: reconstructed images).}
    \label{fig:iDLG_attack}

\end{figure}

In Table~\ref{tab:exp_results2}, we note the ASR values (4th column) for the CPL, DLG, and iDLG attack methods, indicating that approximately 50$\%$, 55$\%$, and 75$\%$ of images have been successfully reconstructed for Melanoma Skin Cancer, COVID-19 X-ray, and Brain Tumor MRI Images, respectively. SSIM and MSE for successful attacks are also shown in the 5th and 6th column in Table \ref{tab:exp_results2}. Additionally, we compare the execution time required for performing the privacy attacks across all scenarios (7th column in Table \ref{tab:exp_results2}). Figure~\ref{fig:Attack_result1}, Figure~\ref{fig:Attack_result2}, and Figure~\ref{fig:Attack_result3} illustrate the original training image and the intermediate reconstructed images resulting from a successful CPL attack on Melanoma Skin Cancer, COVID-19 X-ray, and Brain Tumor MRI Images, respectively, over 200 iterations. We also show some representative medical images and the corresponding reconstructed images after performing iDLG attack~\cite{zhao2020idlg} in Figure \ref{fig:iDLG_attack} from all three datasets. Table~\ref{tab:exp_results2} and these figures demonstrate that patients' private medical images can be precisely reconstructed with minimal noise from the gradients in the FL environment. \vspace{-2ex}

\begin{figure}[h]
    \centering
    \includegraphics[width=1.0\linewidth]{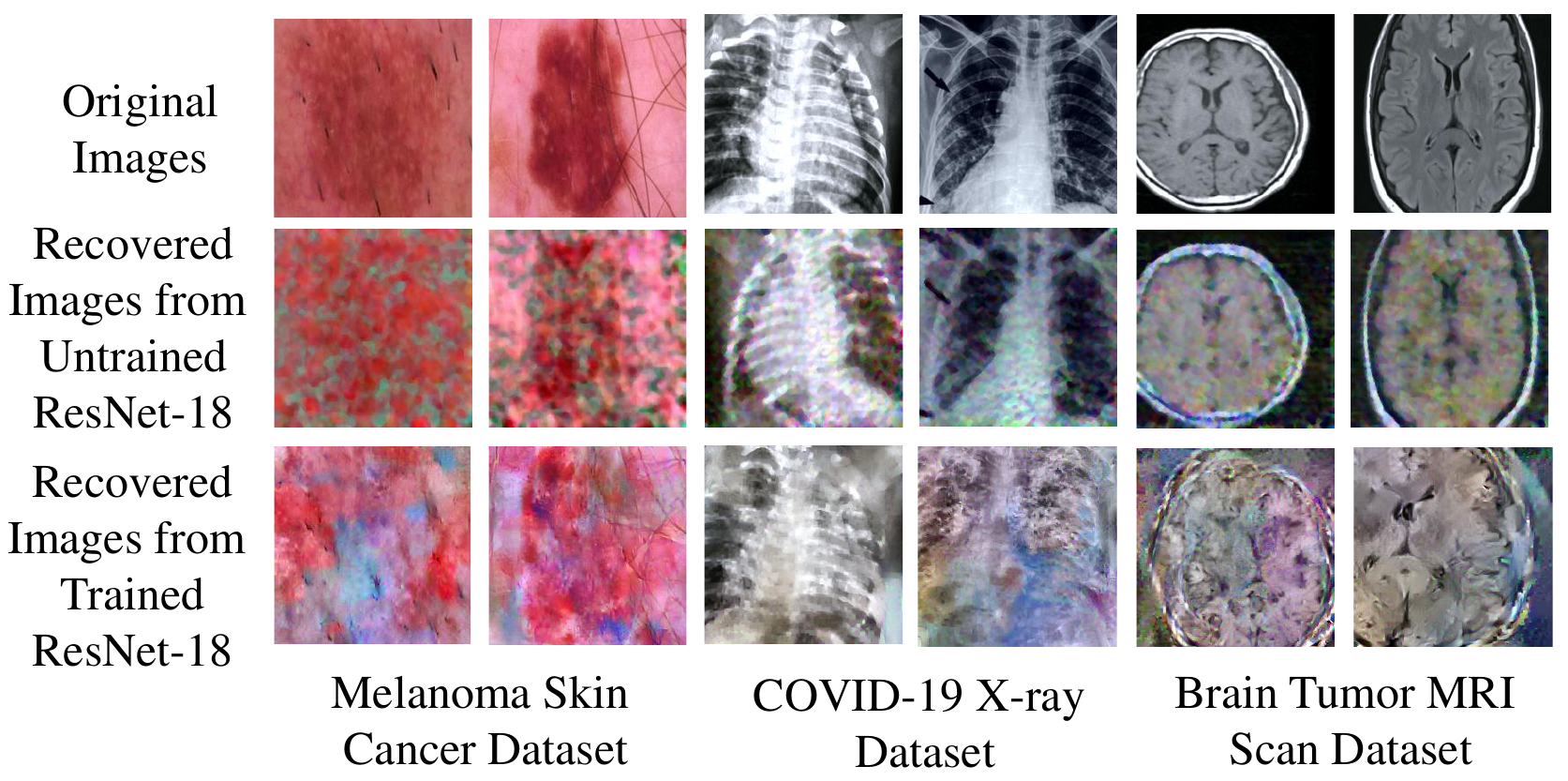}
    \vspace{-4ex}
    \caption{GradInv attack recovered images after 24000 iterations on both untrained and trained ResNet-18 (Top row: original, Middle row: recovered from untrained ResNet-18, Bottom Row: recovered from trained ResNet-18).}
    \label{fig:inv_grad_attack}
    
\end{figure}

We also present reconstructed images after performing GradInv attack on all the datasets for both ResNet-18 \cite{he2016deep} trained and untrained versions with CPL, DLG, and iDLG attack methods. We show the performance of GradInv attack method in the first two rows of Table~\ref{tab:exp_results2} on all corresponding datasets. We observed that the performance of GradInv is better on the untrained ResNet-18 model than the trained ResNet-18 in terms of all evaluation metrics (higher ASR, higher SSIM, and lower MSE). However, the reconstruction process of the GradInv attack for the original private medical image on the untrained ResNet-18 takes a little more time than the trained ResNet-18, which is consistent with the results of GradInv~\cite{geiping2020inverting}. Figure~\ref{fig:inv_grad_attack} visually illustrates the attack performance by the GradInv on Melanoma Skin Cancer, COVID-19 X-ray, and Brain Tumor MRI datasets respectively for 24,000 iterations which supports our observations from Table~\ref{tab:exp_results2}. Comparing the performance of CPL, DLG, and iDLG (in Table \ref{tab:exp_results2}), we observe that GradInv can reconstruct high-quality images by analyzing models with advanced architecture, such as ResNet-18. Though GradInv takes much longer time than CPL, DLG, and iDLG, the performance is better (sometimes similar) in terms of reconstruction quality as well as ASR evaluation metric. \vspace{-1ex}

\subsection{Defense Mechanism Configuration}

In this paper, we perform gradient perturbation as a defense mechanism for preventing privacy leakage attacks. Specifically, we investigate the insertion of a controlled amount of Laplacian noise, characterized by zero mean, into the gradients $\nabla L_k (\omega_t)$  during transmission to the central server, as illustrated in Figure~\ref{fig:attack_defense_mech}(b). The addition of noise to the $\nabla L_k (\omega_t)$ introduces uncertainty in local updates, obscures fine details, and thwarts adversaries' attempts to accurately reconstruct private medical image data. Through our empirical analysis, we find that the default level of noise may not always provide sufficient defense. Therefore, we experiment with varying levels of noise to identify robust defense configurations.

\begin{table}[!h]
\vspace{-2ex}
\caption{CPL attack and defense mechanism performance for different noise levels on three benchmark datasets and CIFAR-10}
\centering
\scalebox{.85}{
\begin{tabular}{|cc|cc|ccc|}
\hline
\multicolumn{2}{|c|}{\multirow{2}{*}{\textbf{Dataset}}} & \multicolumn{2}{c|}{\textbf{CPL Attack}}              & \multicolumn{3}{c|}{\textbf{Defense}}                                                                                                                                                                                                                              \\ \cline{3-7} 
\multicolumn{2}{|c|}{}                                  & \multicolumn{1}{c|}{\textbf{MSE}} & \textbf{SSIM} & \multicolumn{1}{c|}{\textbf{\begin{tabular}[c]{@{}c@{}}Noise \\ Levels\end{tabular}}} & \multicolumn{1}{c|}{\textbf{MSE}}                                                              & \textbf{SSIM}                                                             \\ \hline
\multicolumn{1}{|c|}{1.} & Melanoma Skin Cancer & \multicolumn{1}{c|}{0.0762}      & 0.50        & \multicolumn{1}{c|}{\begin{tabular}[c]{@{}c@{}}100\\ 200\\ 300\\ \textbf{400}\end{tabular}}    & \multicolumn{1}{c|}{\begin{tabular}[c]{@{}c@{}}0.1306\\ 0.1468\\ 0.1497\\ \textbf{0.1503}\end{tabular}} & \begin{tabular}[c]{@{}c@{}}0.0154\\ 0.0131\\ 0.0121\\ \textbf{0.0101}\end{tabular} \\ \hline
\multicolumn{1}{|c|}{2.} & COVID-19 X-ray            & \multicolumn{1}{c|}{0.0641}      & 0.55        & \multicolumn{1}{c|}{\begin{tabular}[c]{@{}c@{}}100\\ 200\\ 300\\ \textbf{400}\end{tabular}}    & \multicolumn{1}{c|}{\begin{tabular}[c]{@{}c@{}}0.0013\\ 0.0157\\ 0.0206\\ \textbf{0.0578}\end{tabular}} & \begin{tabular}[c]{@{}c@{}}0.9815\\ 0.7410\\ 0.7000\\ \textbf{0.4605}\end{tabular} \\ \hline
\multicolumn{1}{|c|}{3.} & Brain Tumor MRI Images        & \multicolumn{1}{c|}{0.0586}      & 0.75        & \multicolumn{1}{c|}{\begin{tabular}[c]{@{}c@{}}100\\ 200\\ 300\\ \textbf{400}\end{tabular}}    & \multicolumn{1}{c|}{\begin{tabular}[c]{@{}c@{}}0.0207\\ 0.0686\\ 0.0300\\ \textbf{0.1705}\end{tabular}} & \begin{tabular}[c]{@{}c@{}}0.6657\\ 0.2383\\ 0.3661\\ \textbf{0.0699}\end{tabular} \\ \hline
\multicolumn{1}{|c|}{4.} & CIFAR-10       & \multicolumn{1}{c|}{0.0222}      & 0.86        & \multicolumn{1}{c|}{\begin{tabular}[c]{@{}c@{}}100\\ 200\\ 300\\ \textbf{400}\end{tabular}}    & \multicolumn{1}{c|}{\begin{tabular}[c]{@{}c@{}}0.0265\\ 0.0292\\ 0.0346\\ \textbf{0.0475}\end{tabular}} & \begin{tabular}[c]{@{}c@{}}0.4965\\ 0.4366\\ 0.3270\\ \textbf{0.2295}\end{tabular} \\ 
\hline
\end{tabular}
}

\label{tab:exp_results3}

\end{table}
\vspace{-1.5ex}\subsection{Performance of Defense Mechanisms.}

\begin{figure}[!hb]
    \centering
    
     \includegraphics[scale=0.33]{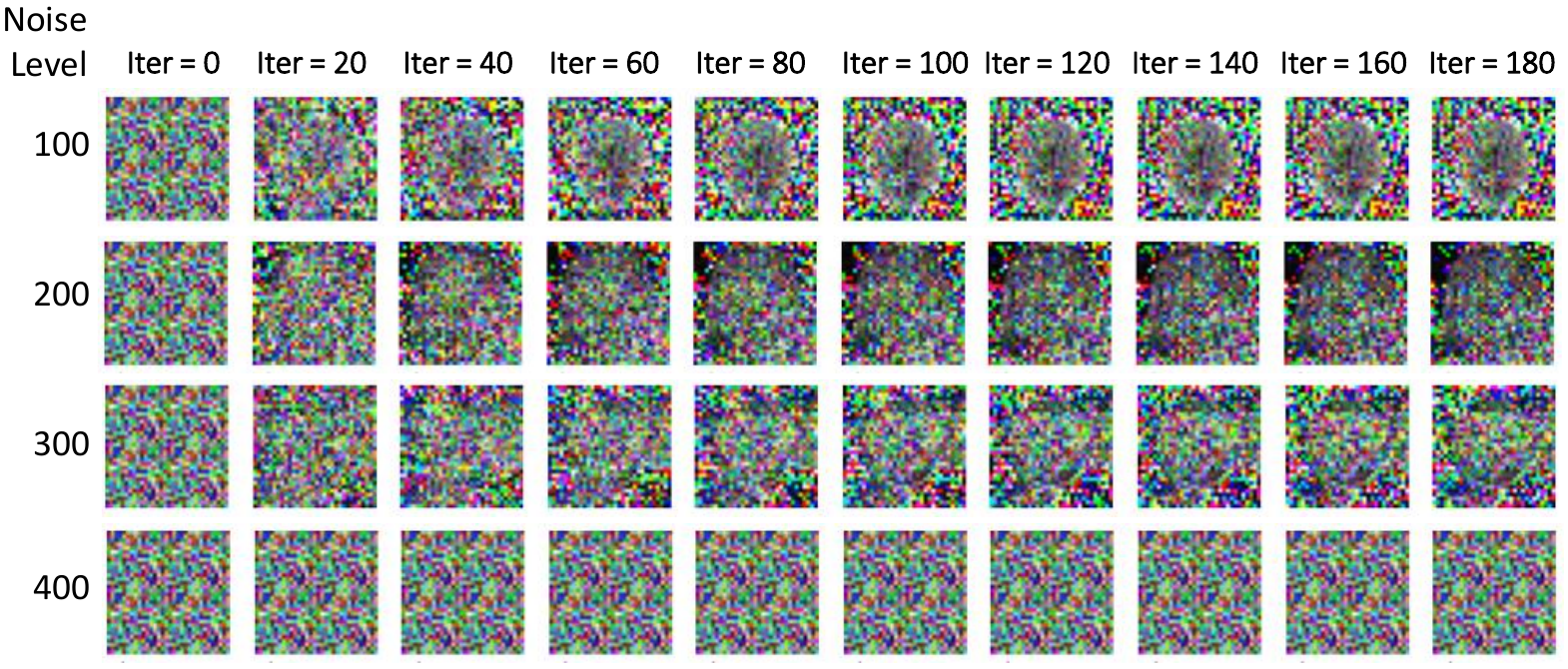}
     
     \caption{Defense to CPL attack on Brain Tumor MRI Dataset}
     \label{fig:defense_result3}
  
\end{figure}

The aim of defense against privacy attacks is to enhance the dissimilarity, measured by metrics such as MSE, and minimize the similarity, quantified by metrics like SSIM, between the original private training images and the reconstructed images by performing privacy attacks. As presented in Table~\ref{tab:exp_results3}, we observe that defense becomes stronger as we enhance the level of Laplacian noise to the gradients as shown in Figure~\ref{fig:attack_defense_mech}(b). Introducing random noise to $\nabla L_k (\omega_t)$ makes the extraction of sensitive information from local gradients more challenging in FL environment. Figure~\ref{fig:defense_result3} visualizes the outcome of the defense mechanism for noise levels at 100, 200, 300, and 400 respectively for the Brain Tumor MRI  dataset. At lower noise levels, such as 100, private information remains susceptible to extraction under the CPL attack, as shown by the first row of Figure~\ref{fig:defense_result3}. This suggests that only adding a standard amount of random noise may not always offer sufficient privacy protection for medical data within FL settings.

Furthermore, we extend our analysis to include a comparison of privacy risks of medical images with a generic dataset, CIFAR-10~\cite{CIFAR10}, and present the corresponding SSIM and MSE values for equivalent noise levels. This comparison outlined in the last row of Table~\ref{tab:exp_results3}, provides insights into the privacy risks between generic datasets and medical datasets. The comparison of SSIM and MSE values for all four noise levels in CIFAR-10 reveals that the standard level of noise applied to gradients may provide sufficient privacy protection for generic images. However, it may not always provide a sufficiently robust defense for medical images.

\begin{table}[!ht]

\caption{Statistical Distributions of Medical Image Datasets and Generic Image Dataset}
\label{tab:stst_dist}

\begin{tabular}{|c|cc|}
\hline
\multirow{2}{*}{\textbf{Datasets}}                                      & \multicolumn{2}{c|}{\textbf{Properties}}                                                                                                                                                                                   \\ \cline{2-3} 
                                                                        & \multicolumn{1}{c|}{\textbf{\begin{tabular}[c]{@{}c@{}}Mean\end{tabular}}} & \textbf{\begin{tabular}[c]{@{}c@{}}Standard  Deviation\end{tabular}} \\ \hline
\begin{tabular}[c]{@{}c@{}}Melanoma Skin \\ Cancer\end{tabular} & \multicolumn{1}{c|}{{[}0.7160, 0.5668, 0.5441{]}}                                                              & {[}0.2207, 0.2087, 0.2222{]}                                                                              \\ \hline
\begin{tabular}[c]{@{}c@{}}COVID-19 \\ X-ray \end{tabular}           & \multicolumn{1}{c|}{{[}0.4949, 0.4950, 0.4953{]}}                                                              & {[}0.2687, 0.2687, 0.2688{]}                                                                              \\ \hline
\begin{tabular}[c]{@{}c@{}}Brain Tumor\\ MRI Scans\end{tabular}         & \multicolumn{1}{c|}{{[}0.1869, 0.1869, 0.1870{]}}                                                              & {[}0.1763, 0.1763, 0.1763{]}                                                                              \\ \hline
CIFAR-10                                                                & \multicolumn{1}{c|}{{[}0.4914, 0.4822, 0.4467{]}}                                                              & {[}0.2471, 0.2434, 0.2615{]}                                                                              \\ \hline
\end{tabular}
\vspace{-2ex}
\end{table}

\section{Discussion}\label{section:discussion}
After performing the experiments for different attack methods and defense mechanisms with various configurations on three medical datasets, we discuss and address several research questions (\textbf{RQ}).

\textbf{RQ1: What are the unique challenges for privacy protection of medical images?} Apart from the complexity, and high dimensionality, as mentioned in Section~\ref{section:motivation}, the statistical distribution of medical images often deviates from the generic images, which implies significant differences between these two data categories in terms of processing, privacy protection, and execution time. From the distribution of the datasets as shown in 3 channels tensor format in Table V, for medical images (first three rows) and the generic image (CIFAR-10)~\cite{CIFAR10}, we can observe that the mean and standard deviation of medical images may differ from generic images.  
\begin{table*}[!ht]
\centering
\caption{Model Performance of trained Resnet-18 on original images, recovered images by CPL attack, and recovered images under  perturbed (Laplacian noise) gradients for three medical datasets and CIFAR-10.}

\scalebox{.98}
{
\begin{tabular}{|cc|cccc|}
\hline

\multicolumn{2}{|c|}{}                                                                                                            & \multicolumn{4}{c|}{\textbf{Accuracy}}                                                                                                                                                                                                                                                                                 \\ \hline
\multicolumn{2}{|c|}{\textbf{\begin{tabular}[c]{@{}c@{}}Data Source\end{tabular}}}                                              & \multicolumn{1}{c|}{\textbf{CIFAR-10}} & \multicolumn{1}{c|}{\textbf{\begin{tabular}[c]{@{}c@{}}Melanoma Skin \\ Cancer \end{tabular}}} & \multicolumn{1}{c|}{\textbf{\begin{tabular}[c]{@{}c@{}}COVID-19\\ X-ray\end{tabular}}} & \textbf{\begin{tabular}[c]{@{}c@{}}Brain Tumor\\ MRI Scans\end{tabular}} \\ \hline
\multicolumn{2}{|c|}{\begin{tabular}[c]{@{}c@{}}Original Data\end{tabular}}                                                     & \multicolumn{1}{c|}{0.82}              & \multicolumn{1}{c|}{0.97}                                                                             & \multicolumn{1}{c|}{0.95}                                                                  & 0.96                                                                     \\ \hline
\multicolumn{2}{|c|}{\begin{tabular}[c]{@{}c@{}}Recovered Data\end{tabular}}                                                    & \multicolumn{1}{c|}{0.74}              & \multicolumn{1}{c|}{0.95}                                                                             & \multicolumn{1}{c|}{0.87}                                                                  & 0.74                                                                     \\ \hline
\multicolumn{1}{|c|}{\multirow{2}{*}{\begin{tabular}[c]{@{}c@{}}Perturbed gradients with\\ Different Noise Levels\end{tabular}}} & 100 & \multicolumn{1}{c|}{0.12}              & \multicolumn{1}{c|}{\textbf{0.96}}                                                                             & \multicolumn{1}{c|}{\textbf{0.90}}                                                                  & \textbf{0.92}                                                                     \\ \cline{2-6} 
\multicolumn{1}{|c|}{}                                                                                                      & 400 & \multicolumn{1}{c|}{0.09}              & \multicolumn{1}{c|}{\textbf{0.93}}                                                                             & \multicolumn{1}{c|}{\textbf{0.88}}                                                                  & \textbf{0.87}                                                                     \\ \hline
\end{tabular}

}

\label{tab:performance_comp}

\end{table*}

\begin{figure}[h]
    \centering

    \includegraphics[scale=0.33]{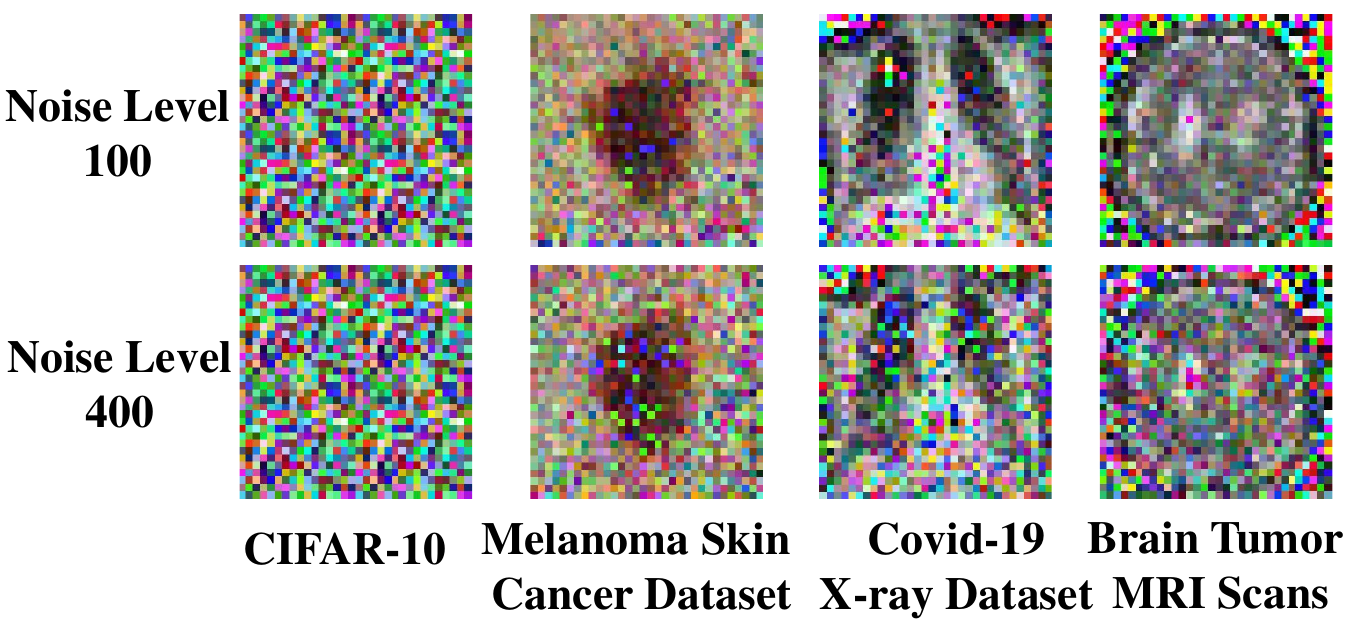}
  
    \caption{Dataset Comparison Under Different Noise Levels}
    \label{fig:performance_comp}

\end{figure}

\vspace{2ex}\textbf{RQ2: Which level of Laplacian noise is enough to safeguard the privacy of medical images?} As shown and discussed in the previous section, medical image datasets (Brain Tumor MRI dataset,  Figure \ref{fig:defense_result3}) are still susceptible to being revealed even if we add the highest level of noise compared with a generic image dataset (e.g., CIFAR-10). We show several samples in Figure~\ref{fig:performance_comp} after adding the lowest and highest levels of Laplacian noise to the generic CIFAR-10 image and the medical images that we studied in this research respectively. From these visual examples, it can be clearly observed that the lowest level of noise (100) is enough to protect the content of CIFAR-10 images. On the other hand, by looking at the medical images of all datasets, even if we add the highest level of noise (400), a major portion of the contents is still visible, which may reveal their categories. Thus, it requires further studies to build a strong defense for medical images.   

\textbf{RQ3: If we keep increasing the noise level to make a stronger defense against privacy leakage attacks, how does it impact the model performance?} In order to check the model performance on the recovered images by CPL attack and the reconstructed images after perturbing gradients with different noise levels, we employed trained ResNet-18 to evaluate image classification accuracy on the original data, recovered data, and recovered data under perturbed gradients as shown in Table \ref{tab:performance_comp}. We observe a slight performance drop for reconstructed images by CPL, and a significant drop occurred for the recovered data under perturbed gradients at any noise levels in terms of classification accuracy for the generic image dataset (CIFAR-10). On the contrary, for all medical image datasets under the perturbed gradients, although we observed a slight performance drop in the classification accuracy, it still remains high. The reason behind such high accuracy is that even if we added the highest level of noise to the gradients, most of the contents were visible for all medical images (see medical images in Figure~\ref{fig:performance_comp}). 
This further confirms the unique research challenges associated with privacy protection for medical images.

\section{Conclusion} \label{section:conclusion} 
This paper introduces MedPFL, a framework designed to facilitate the analysis and mitigation of privacy risks associated with medical images in the FL environment. We demonstrate the substantial privacy risks inherent in utilizing FL for medical data processing, where sensitive patient data can be susceptible to recovery by adversaries through various privacy attacks. In our study, we employ different levels of random noise as a defense mechanism against these privacy attacks. However, we observe that while higher levels of noise can offer stronger privacy protection, however, adding random noise may not always adequately safeguard medical images within FL environments. Through the experiments of real-world scenarios involving multiple privacy attacks on medical images across three benchmark datasets, we underscore the critical challenges associated with mitigating privacy risks within FL, particularly within the medical domain. In the future, we intend to explore other types of privacy attacks and devise innovative privacy-preserving techniques tailored specifically to safeguarding medical data within FL settings. Also, we plan to incorporate different learning tasks, e.g., medical image segmentation into the MedPFL framework.

\vspace{-1ex}
\bibliographystyle{ieeetr}
\bibliography{ref}
\vspace{-3ex}

\newpage

\begin{IEEEbiography}
[{\includegraphics[width=1in,height=1.25in,clip,keepaspectratio]{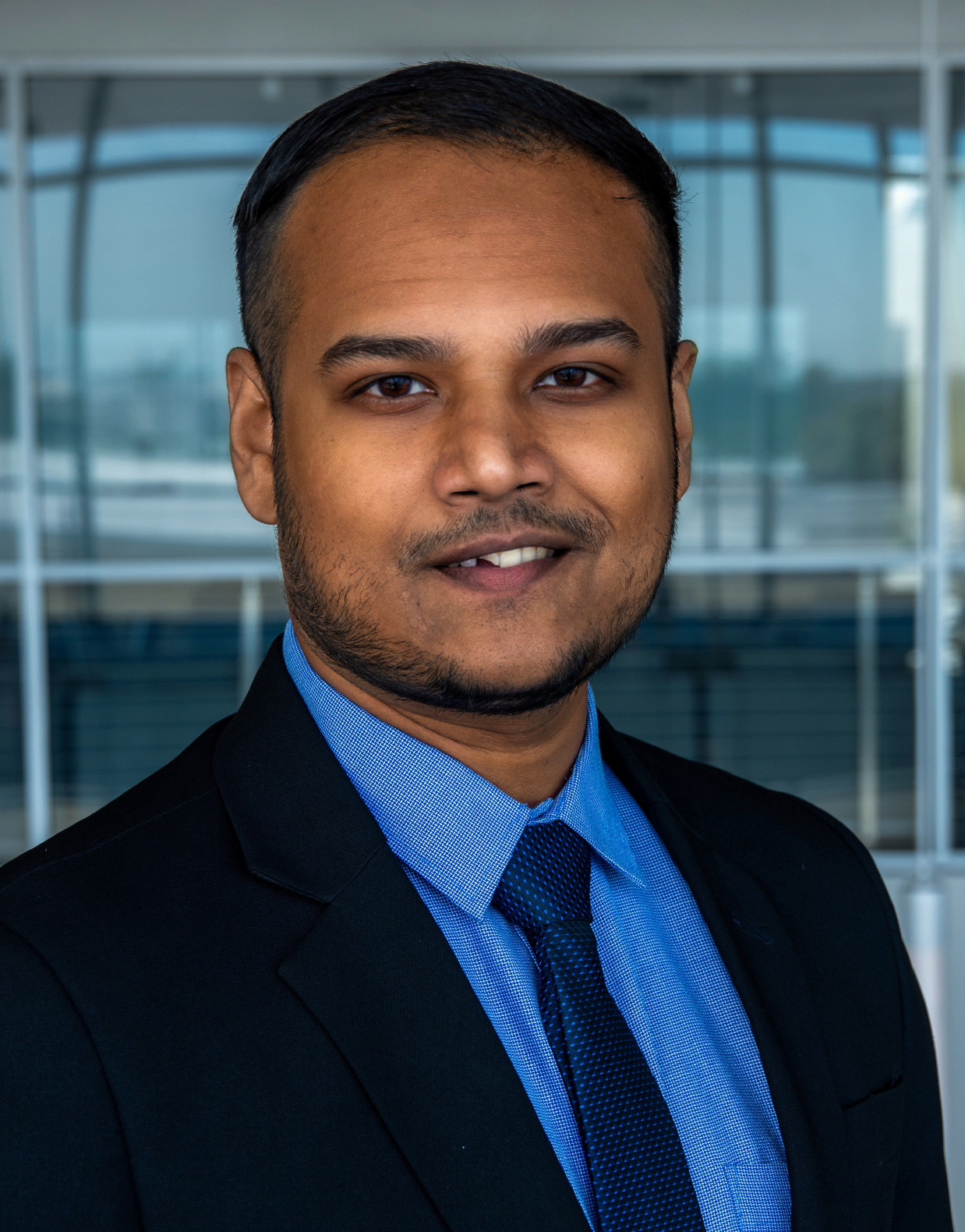}}]{Badhan Chandra Das} is a Ph.D. Candidate at Knight Foundation School of Computing and Information Sciences (KFSCIS), Florida International University (FIU). His research includes privacy-preserving machine learning algorithms, computer vision, and their real-world applications. 
\end{IEEEbiography}

\begin{IEEEbiography}
[{\includegraphics[width=1in,height=1.25in,clip,keepaspectratio]{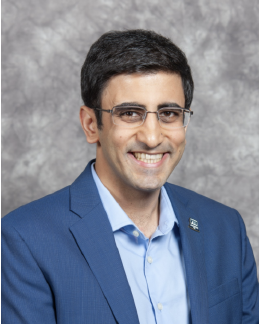}}]{M. Hadi Amini, Senior Member, IEEE} is an Assistant Professor at Knight Foundation School of Computing and Information Sciences at Florida International University. He is the director of Sustainability, Optimization, and Learning for InterDependent networks laboratory (www.solidlab.network). He received his Ph.D. in Electrical and Computer Engineering from Carnegie Mellon University in 2019, where he received his M.Sc. degree in 2015. He also holds a doctoral degree in Computer Science and Technology. He serves as the Director and PI of  ADvanced education and research for Machine learning-driven critical Infrastructure REsilience (ADMIRE) Center, Supported by the U.S. DHS; and 
Associate Director of the National Center for Transportation Cybersecurity and Resiliency (TraCR), Supported by the U.S. DOT. He is an Associate Editor of \textit{IEEE Transactions on Information Forensics and Security}. His research interests include secure and privacy-preserving distributed optimization and learning algorithms, interdependent networks, and cyber-physical-social security and resilience. 
\end{IEEEbiography}

\begin{IEEEbiography}[{\includegraphics[width=1in,height=1.25in,clip,keepaspectratio]{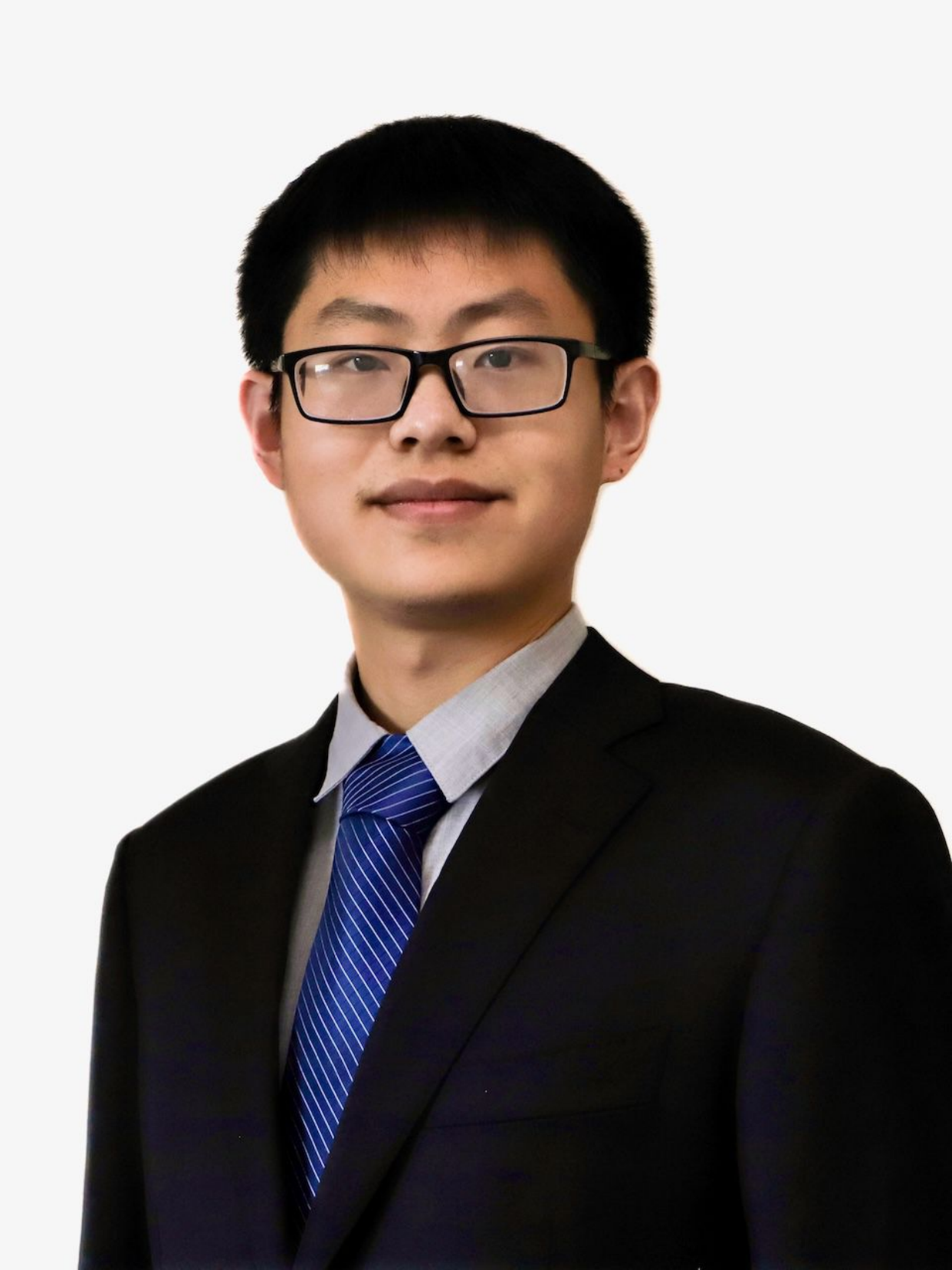}}]{Yanzhao Wu} is an Assistant Professor in the Knight Foundation School of Computing and Information Sciences (KFSCIS) at Florida International University (FIU). He obtained his Bachelor’s degree from University of Science and Technology of China (USTC) in 2017 and then received his PhD in Computer Science from Georgia Institute of Technology in 2022. His research interests are primarily centered on the intersection of machine learning and computing systems, including machine learning algorithm and system co-design, large language models (LLMs), edge AI, privacy-preserving machine learning, deep learning, big data analytics, and their real-world applications.
\end{IEEEbiography}

\vfill

\end{document}